\documentclass[conference]{IEEEtran}
\IEEEoverridecommandlockouts
\usepackage{cite}
\usepackage{amsmath,amssymb,amsfonts}
\usepackage{algorithmic}
\usepackage{graphicx}
\usepackage{textcomp}
\usepackage{xcolor}
\def\BibTeX{{\rm B\kern-.05em{\sc i\kern-.025em b}\kern-.08em
    T\kern-.1667em\lower.7ex\hbox{E}\kern-.125emX}}

\usepackage{url}
\usepackage{bbm}

\usepackage{fancyhdr}

\usepackage[hidelinks]{hyperref}
\hypersetup{
  colorlinks   = true, 
  urlcolor     = blue, 
  citecolor   = green 
}

\begin{document}
\title{LG-Hand: Advancing 3D Hand Pose Estimation with Locally and Globally Kinematic Knowledge*\\
\thanks{This research is funded by Hanoi University of Science and Technology (HUST) under project number T2021-SAHEP-003.}}

\author{\IEEEauthorblockN{Tu Le-Xuan\IEEEauthorrefmark{1}\textsuperscript{\textsection},
Trung Tran-Quang\IEEEauthorrefmark{3}\textsuperscript{\textsection}, Thi Ngoc Hien Doan\IEEEauthorrefmark{1}\IEEEauthorrefmark{2} and
Thanh-Hai Tran\IEEEauthorrefmark{1}\IEEEauthorrefmark{2}}
\\
\IEEEauthorblockA{\IEEEauthorrefmark{1}School of Electrical and Electronic Engineering (SEEE), Hanoi University of Science and Technology, Hanoi, Vietnam \\
\IEEEauthorrefmark{2}MICA International Research Institute, Hanoi University of Science and Technology, Hanoi, Vietnam \\
\IEEEauthorrefmark{3}Research and Development Team, Asilla Inc., Tokyo, Japan \\
Email: \textit{letu071299@gmail.com}, \textit{tranquangtrunghnvn@gmail.com}, \textit{\{hien.doanthingoc,hai.tranthithanh1\}@hust.edu.vn}
}
}
\maketitle
\begingroup\renewcommand\thefootnote{\textsection}
\footnotetext{Equal contribution. This research was done when Tu was an intern at Research and Development Team, Asilla Inc.}
\endgroup

\thispagestyle{fancy}
\pagestyle{fancy}
\fancyhf{}
\lhead{The 14th IEEE International Conference on Knowledge and Systems Engineering (KSE 2022)}
\rhead{October 19-21, 2022}
\cfoot{\thepage}

\begin{abstract}
3D hand pose estimation from RGB images suffers from the difficulty of obtaining the depth information. Therefore, a great deal of attention has been spent on estimating 3D hand pose from 2D hand joints. In this paper, we leverage the advantage of spatial-temporal Graph Convolutional Neural Networks and propose LG-Hand, a powerful method for 3D hand pose estimation. Our method incorporates both spatial and temporal dependencies into a single process. We argue that kinematic information plays an important role, contributing to the performance of 3D hand pose estimation. We thereby introduce two new objective functions, Angle and Direction loss, to take the hand structure into account. While Angle loss covers locally kinematic information, Direction loss handles globally kinematic one. Our LG-Hand achieves promising results on the First-Person Hand Action Benchmark (FPHAB) dataset. We also perform an ablation study to show the efficacy of the two proposed objective functions.
\end{abstract}

\begin{IEEEkeywords}
hand pose estimation, kinematic knowledge, spatial-temporal graph, local and global constraints
\end{IEEEkeywords}

\section{Introduction}

Hand is one of vital parts of human, serving a wide range of actions such as holding, writing, or handshaking. Hand pose estimation, an interesting task of computer vision, thereby plays a critical role in real-world applications. These applications include robotics, human-machine interaction, and entertainment. When it comes from 2D to 3D prediction, both hand and human pose estimation suffer from depth ambiguity \cite{li2019generating} and multi-view variation \cite{dong2019fast}. The hand, a deformable object with a high degree of freedom, is also prune to be occluded as it interacts with the world, so 3D hand pose estimation still remains challenging.

Most of existing methods try to solve 3D hand pose estimation by firstly localizing joints in a 2D space, then lifting 3D joints from estimated 2D joints \cite{garcia2018first,ge2016robust,panteleris2018using,doosti2020hope,cai2019exploiting}. The depth information can be used for 3D hand joints prediction, as shown in \cite{garcia2018first,ge2016robust}. Especially, Liuhao et al. \cite{ge2016robust} utilize the depth image to project 3D points onto three orthogonal planes and then regresses multi-view 2D heatmaps. These heatmaps are finally fused together to estimate 3D hand joint coordinates. However, the depth information is limited in outdoor scenarios and even hard to obtain. Therefore, many efforts have been paid to estimate 3D hand pose from a single RGB image, as proposed in \cite{panteleris2018using,doosti2020hope,cai2019exploiting}. These three methods use off-the-shelf techniques for 2D hand pose estimation and then build 3D estimators to predict 3D hand joints without using the depth information. While Paschalis et al. \cite{panteleris2018using} use non-linear least-squares minimization to fit a 3D hand model to the estimated 2D joints, HOPE-Net \cite{doosti2020hope} and Yujun et al. \cite{cai2019exploiting} take the advantage of Graph Convolutional Neural Networks to reconstruct 3D hand joints from 2D coordinates. Instead of using a spatial graph as in \cite{doosti2020hope}, Yujun et al. \cite{cai2019exploiting} propose exploiting spatial-temporal Graph Convolutional Neural Networks to alleviate the issue of depth ambiguity. However, geometric constraints of hand structure (bone length for example) are not taken into account. Therefore, SST-GCN \cite{le2021sst} introduces a new partial loss function, the consistency of the finger bone length, to better estimate 3D hand joints.

\begin{figure}[h!]
\centerline{\includegraphics[width=0.8\linewidth]{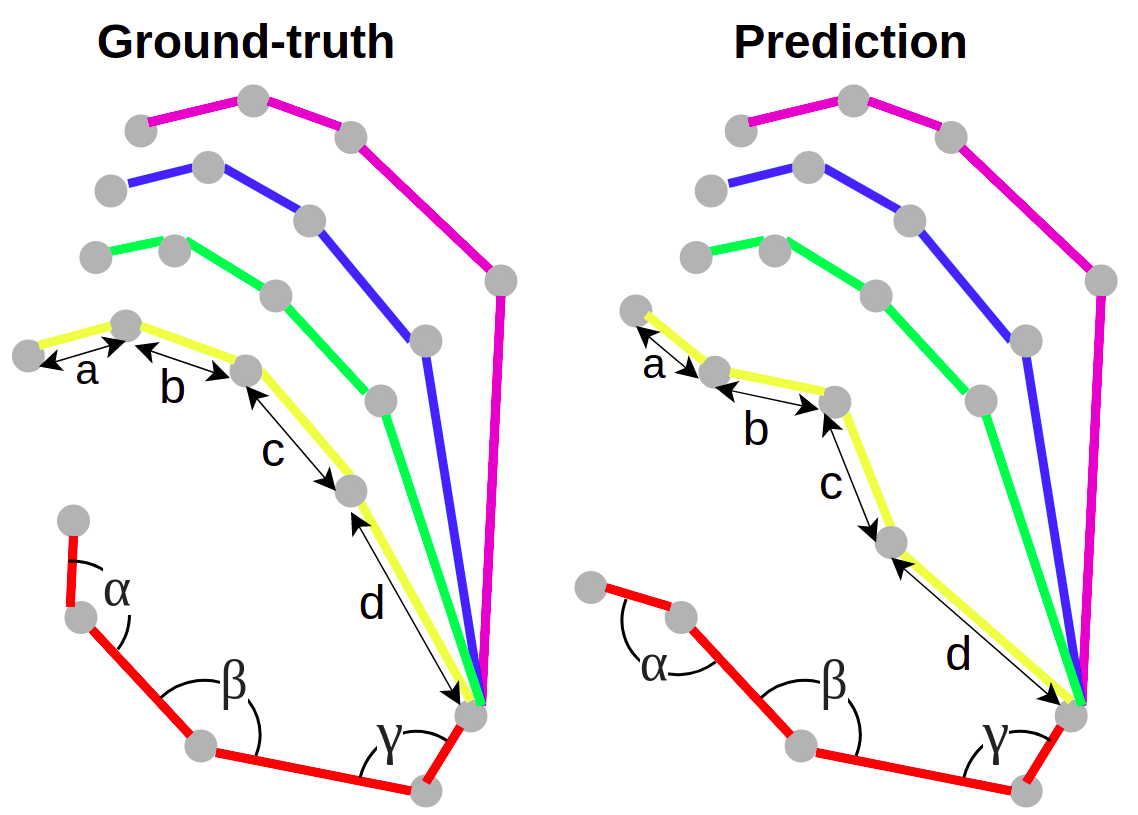}}
\caption{Example of weird predictions even when the finger bone length is the same (the yellow finger) or the angles of the knuckles are matched (the red finger).}
\label{fig_weird_hand}
\end{figure}

\begin{figure*}[h!]
\centerline{\includegraphics[width=1.0\linewidth]{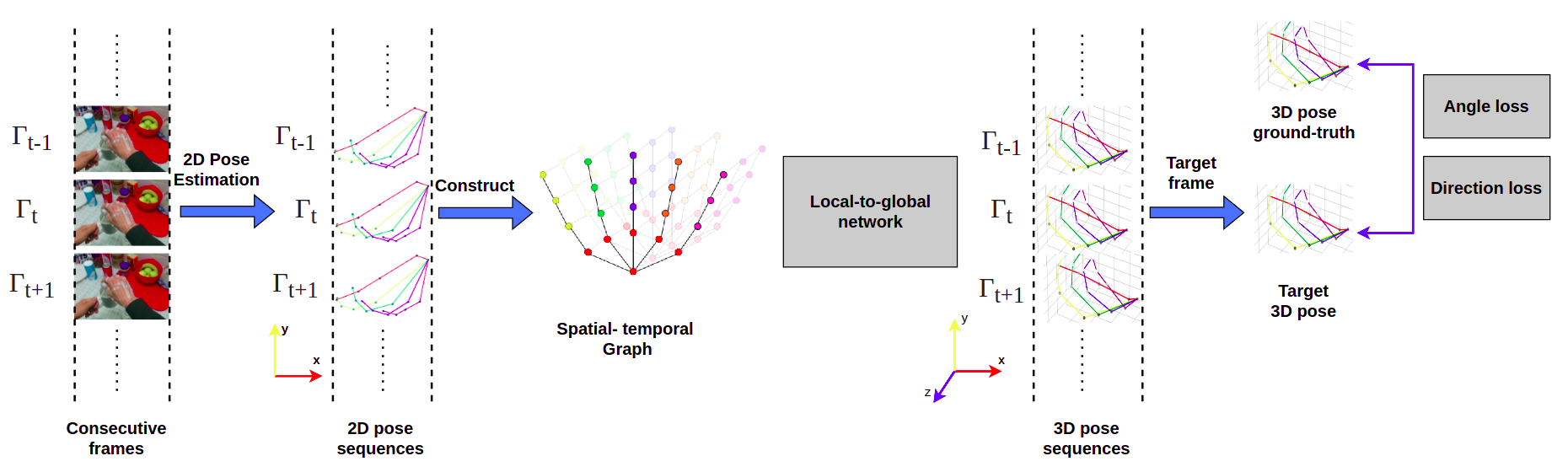}}
\caption{Overall framework of LG-Hand. Our method takes as input a sequence of 2D hand skeletons and outputs 3D hand joint coordinates.}
\label{fig_overall_framework}
\end{figure*}

We argue that the consistency of the finger bone length is not enough for supporting 3D hand pose estimation. For example, when the ground-truth and predicted finger have the same bone length, we are still not sure that the predicted finger matches the ground-truth one. The predicted finger might also look weird, as illustrated by the yellow finger in Fig. \ref{fig_weird_hand}. Motivated by this observation, we propose LG-Hand, a method based on spatial-temporal Graph Convolutional Neural Networks, for 3D hand pose estimation. We introduce locally and globally kinematic constraints to make the predicted 3D hand joints more accurate. Concretely, we encourage the model to learn the angle of the knuckles within a finger, which could be referred to as local information. However, two fingers can still be different even if they have the same angles of the knuckles, as depicted by the red finger in Fig. \ref{fig_weird_hand}. To solve this issue, we thereby add the constraints of the direction of the knuckles, which could be regarded as global information. Our key contributions are summarized as follows:
\begin{itemize}
    \item We propose LG-Hand, an end-to-end framework to estimate 3D hand joints from 2D hand joints.
    \item We introduce new objective functions to model locally and globally kinematic constraints.
    \item We obtain promising results on FPHAB dataset, including the average MPJPE of 17.25 (mm).
\end{itemize}

\section{Related Work}

3D hand joints prediction could be obtained by using magnetic sensors or single RGB images. This section presents the methods for 3D hand pose estimation using RGB images. Especially, Graph Convolutional Neural (GCN) Networks are also reviewed because they directly relate to our method.

\textit{\textbf{3D hand pose estimation from 2D images.}} HOPE-Net \cite{doosti2020hope} utilized off-the-shell 2D pose estimators to generate 2D joints and then introduced Adaptive Graph U-Net architecture to match the hand from 2D to 3D space. Liuhao et al. developed a GCN-based method to generate 3D hand mesh from a single RGB image \cite{ge20193d}. A dataset with 3D annotations was created to train the model in a suppervised manner. The model was then fine-tuned on real-world datasets without 3D annotations. Especially, 3D depth maps were generated from rendered 3D meshes to monitor the training process.

\textit{\textbf{3D hand pose estimation from 2D joints.}} Julieta et al. \cite{martinez2017simple} directly obtained 3D hand joints from 2D coordinates by using conventional components such as linear layers, batch normalization, dropout, and Relu activations. On the other hand, Christian and Thomas \cite{zimmermann2017learning} first generated score maps with respect to hand joint locations. They then used a PosePrior network to predict 3D hand joint coordinates.

\textit{\textbf{Graph Convolutional Neural Networks.}} While conventional convolutions only handle the grid-based data such as images, there are various types of data in reality. These types of data include social data or user interaction. Conventional convolutions showed the limitation in addressing non-grid data. Therefore, GCN Networks were proposed to deal with this problem. We also utilize GCN Networks in our method, which will be presented in Section \ref{sec_method}.

\section{Methodology}
\label{sec_method}

This section first presents the overall framework of LG-Hand especially the procedue of constructing an input hand sequence. Next, we present Graph Convolutional Neural (GCN) Networks to process input hand sequences. We then describe the local-to-global network, one of important parts of our method for multi-scale training. Finally, the objective function, including our locally and globally kinematic constraints, will be shown in detail. Note that we call the method, proposed by Yujun et al. \cite{cai2019exploiting}, ST-GCN for a convenient explanation.  

\subsection{Overall Framework}

\begin{figure*}[h!]
\centerline{\includegraphics[width=1.0\linewidth]{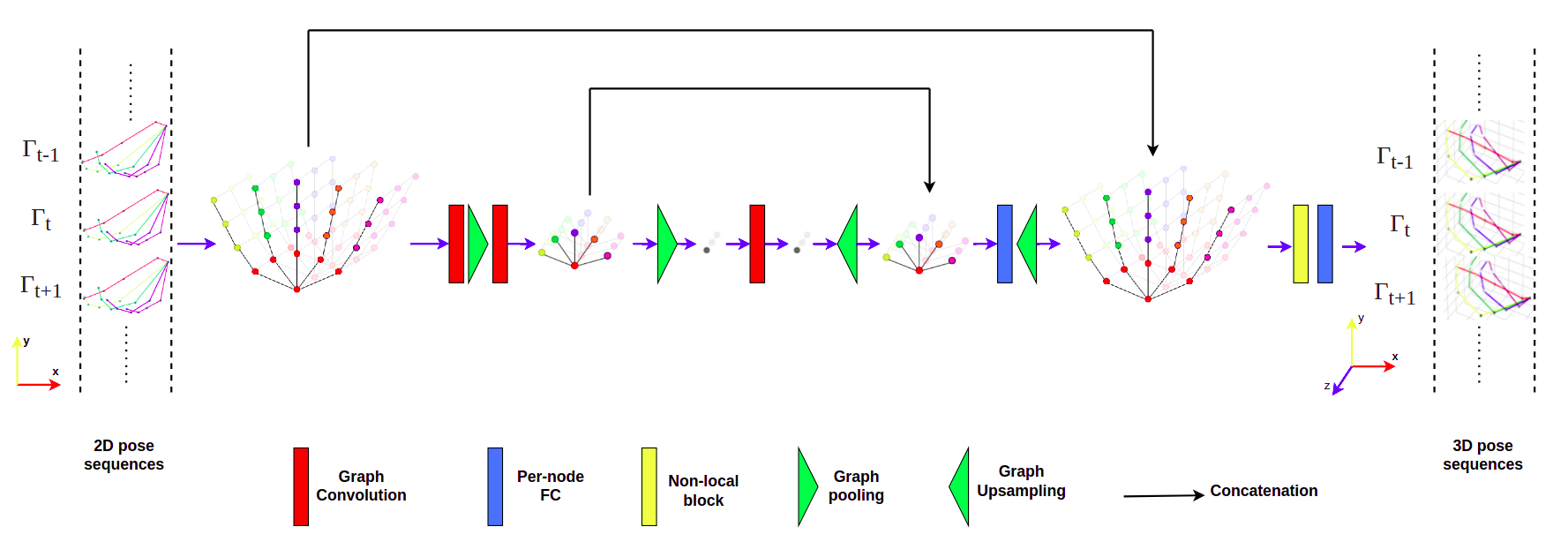}}
\caption{The architecture of GCN-based local-to-global network.}
\label{fig_local_to_global_network}
\end{figure*}

The overall framework of LG-Hand is illustrated in Fig. \ref{fig_overall_framework}. LG-Hand takes as input a sequence of 2D hand skeletons of consecutive frames and outputs the 3D hand joints of the target frame. Concretely, the predicted 2D hand joints of consecutive frames (..., $\Gamma_{t-2}$, $\Gamma_{t-1}$, $\Gamma_t$, $\Gamma_{t+1}$, $\Gamma_{t+2}$, ...) are combined into a spatial-temporal graph and then fed into a GCN-based local-to-global network, which is described in Fig. \ref{fig_local_to_global_network}, to produce 3D hand joints of the frame $\Gamma_t$. At the end of the pipeline, we propose two new objective functions, called Angle and Direction loss, to leverage kinematic characteristics. The GCN-based local-to-global network and the objective function will be elaborated in Section \ref{sec_local_to_global_net} and \ref{sec_objective_function}, respectively. The remainder of this section presents the way to construct the spatial-temporal graph from the sequence of 2D hand skeletons.

\begin{figure}[h!]
\centerline{\includegraphics[width=0.65\linewidth]{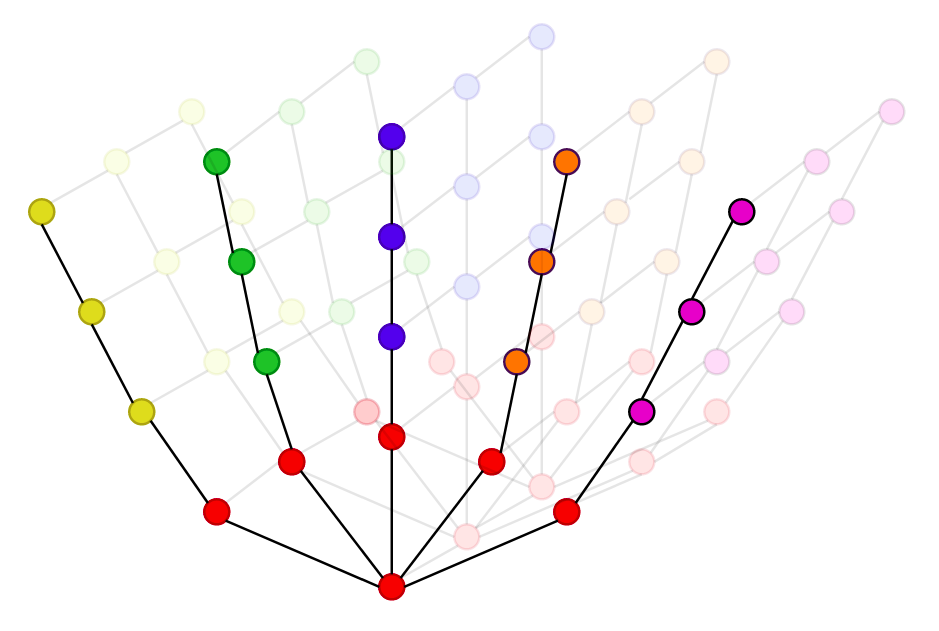}}
\caption{Illustration of spatial-temporal graph.}
\label{fig_st_graph}
\end{figure}

From a sequence of skeletons with $N$ joints and $T$ frames, we construct an undirected spatial-temporal graph $G=(V,E,A)$ as in Fig. \ref{fig_st_graph}, where $V$ and $E$ are the set of nodes and edges of the graph respectively, and $A$ denotes the adjacency matrix. To build the graph $G$, we first connect the joints of one frame following the natural connections of the human hand. Next, the joints of the same type are connected between the consecutive frames. By doing so, our method can work with an arbitraty amount of joints. The set of nodes contains all joints of all input frames: $V=\{v_{ti}: t\in (1,..,T), i\in (1,...,N)\}$. The set of edges is divided into two subsets, $E_p=\{v_{ti}v_{tj}\}$ and $E_s=\{v_{ti}v_{(t+1)i}\}$, for the natural connections within one frame and the connections of the same joints between consecutive frames, respectively. Note that $t$ depicts the frame index, $i$ and $j$ are for the joint index. Finally, the adjacency matrix $A$ is defined as: $A=(a_{ij})_{M\times M}$ with $M=N\times T$. $a_{ij}$ is equal to $0$ if the connection $(i,j)$ is not in $E$. Otherwise, $a_{ij}$ is equal to $1$. 

\subsection{Graph Convolutional Neural Networks for 3D hand pose estimation}

For Graph Convolutional Neural (GCN) Networks, Kipf and Welling \cite{kipf2016semi} proposed:
\begin{equation}
    Z=\widetilde{D}^{-\frac{1}{2}}\widetilde{A}\widetilde{D}^{-\frac{1}{2}}X\Theta\label{equation_graph}
\end{equation}
where $X\in\mathbb{R}^{N\times C}$ is the input signal representing $C$-dimensional features of $N$ nodes on the graph, $\Theta\in\mathbb{R}^{C\times F}$ is the matrix of filter parameters with $F$ is the number of filters, $\widetilde{A}$ and $\widetilde{D}$ are the normalized versions of the adjacency matrix and degree matrix respectively ($\widetilde{A}=A+I_N$, $\widetilde{D}^{ii}=\sum_{j}\widetilde{A}^{ij}$, and $I_N$ is the identity matrix), and $Z\in\mathbb{R}^{N\times F}$ is the convolved signal matrix.

\begin{figure}[h!]
\centerline{\includegraphics[width=0.75\linewidth]{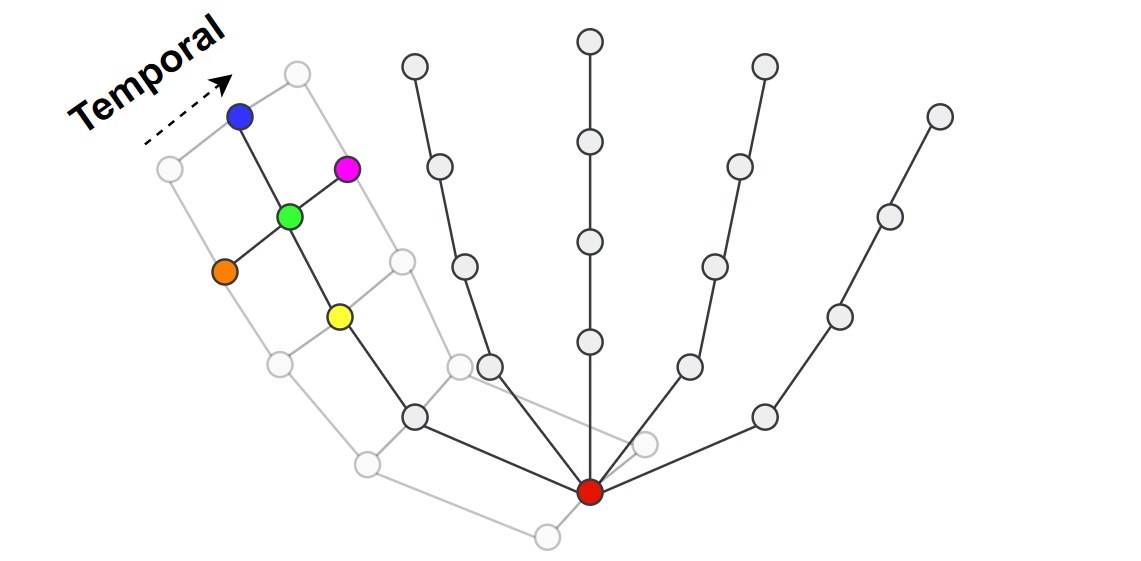}}
\caption{Illustration of neighboring nodes.}
\label{fig_node_classification}
\end{figure}

Equation (\ref{equation_graph}) can be extended for hand pose estimation. First, the neighboring nodes are divided into five groups according to their position relative to the central node, as illustrated in Fig. \ref{fig_node_classification}. These five groups include a central node (green), a time-forward node (orange), a time-backward node (purple), physically-connected nodes containing the one closer (yellow) to and the one further (blue) from the skeleton root (red). GCN Networks are thereby extended to:
\begin{equation}
    Z=\sum_k D_k^{-\frac{1}{2}}A_k D_k^{-\frac{1}{2}}X\Theta_k
\end{equation}
where $k$ is the index of the neighbor types, and $\Theta_k$ is the filter matrix for the $k$-th type with 1-hop neighboring nodes. Notably, the normalized adjacency matrix $\widetilde{A}$ is dismantled into $k$ sub-matrices with $\widetilde{A}=\sum_k A_k$ and $D_k^{ii}=\sum_j A_k^{ij}$.

\subsection{GCN-Based Local-to-Global Network}
\label{sec_local_to_global_net}

GCN-based local-to-global network plays an important role in our method, being responsible for processing and combining the features at different scales, as described in Fig. \ref{fig_local_to_global_network}. The pipeline can be split into two stages: bottom-up and top-down. At the bottom-up stage, the network takes as input the spatial-temporal graph and uses graph convolutional and graph pooling layers to extract the features. The later top-down stage conducts upsampling and combining the upsampled features with the high-resolution ones from the bottom layers. To effectively remain the information learned in the bottom-up stage, an element-wise concatenation is applied for the same-scale features of the bottom-up and top-down stage. Finally, a non-local block is used before generating 3D hand pose sequences to support the full hand reconstruction.

\begin{figure}[h!]
\centerline{\includegraphics[width=0.9\linewidth]{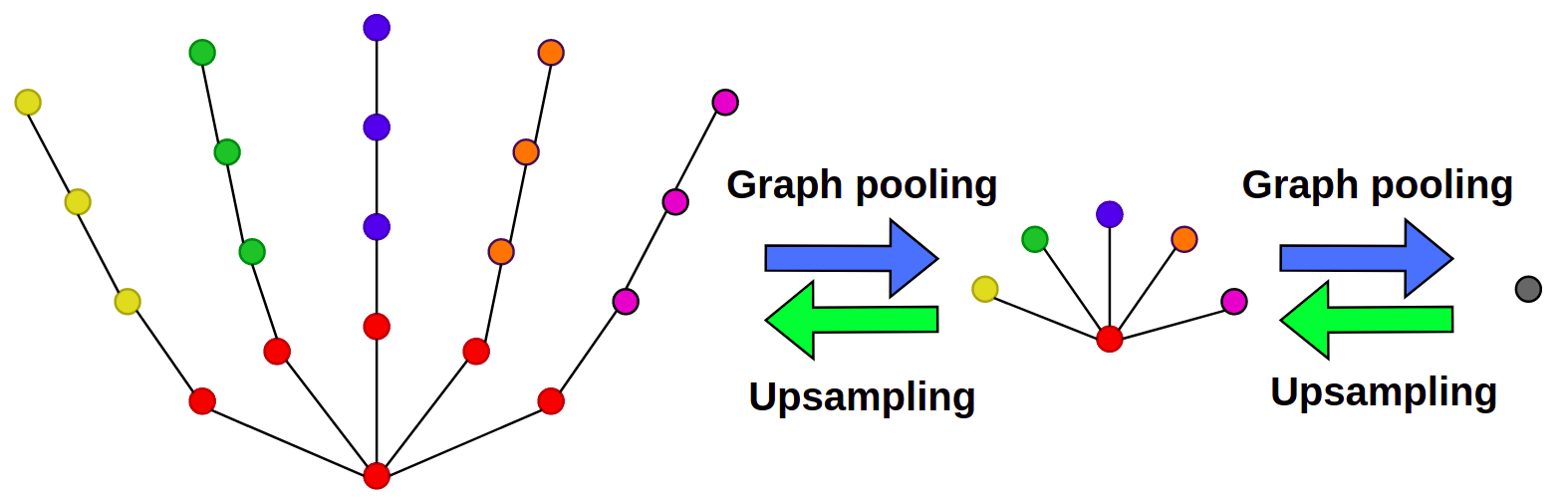}}
\caption{Illustration of graph pooling and upsampling for hand pose estimation. The same color depicts the nodes from the same group.}
\label{fig_graph_pooling}
\end{figure}

One of the important parts of GCN-based local-to-global network is graph pooling and upsampling, as illustrated in Fig. \ref{fig_graph_pooling}. The 21 hand joints are first split into groups according to the palm and the fingers. A max pooling operation is then applied to each group, resulting in a smaller graph with 6 nodes, where each node represents the local information of a region of the hand. An extra max pooling operation is applied to the 6-node graph, leading to a 1-node graph. This 1-node graph contains the global information of the hand. In contrast, upsampling operation is responsible for converting a sparse graph to a full graph, as described in Fig. \ref{fig_graph_pooling}.

\subsection{Objective Function}
\label{sec_objective_function}

\begin{figure}[h!]
\centerline{\includegraphics[width=1.0\linewidth]{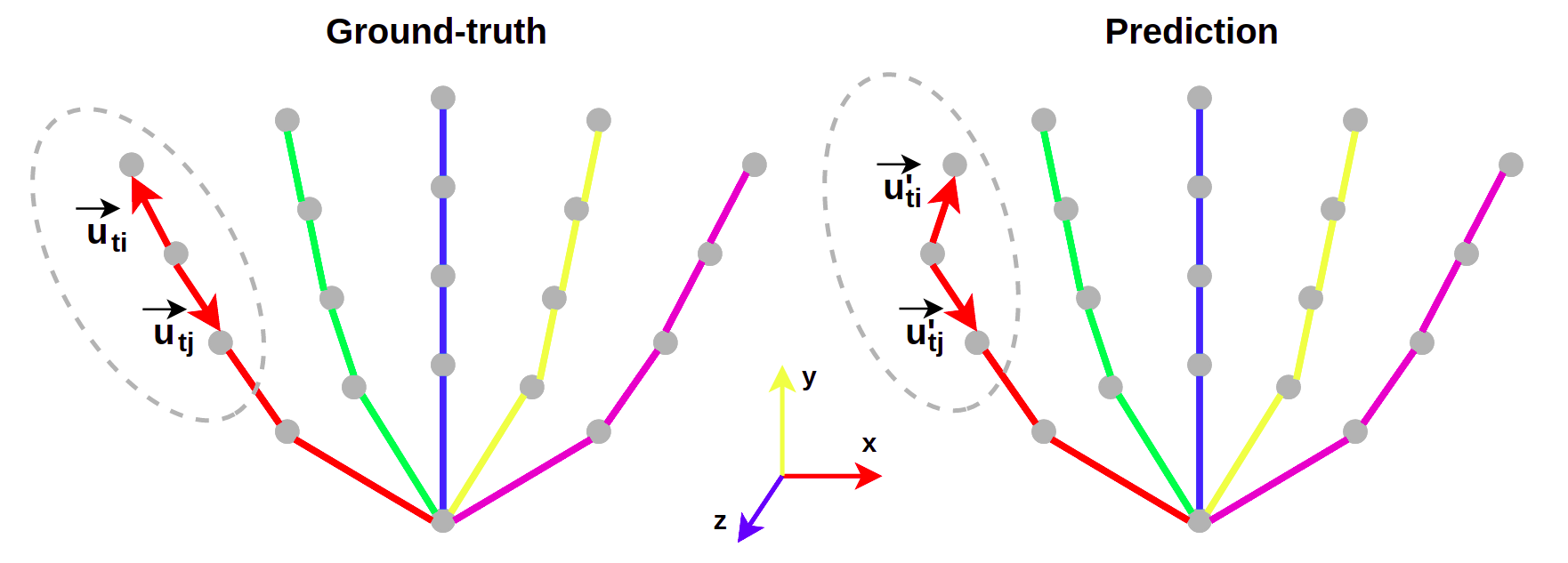}}
\caption{Illustration of vectors for computing Angle and Direction loss.}
\label{fig_angle_direction_loss}
\end{figure}

ST-GCN \cite{cai2019exploiting} and SST-GCN \cite{le2021sst} both utilize spatial-temporal Graph Convolutional Neural Networks for 3D hand pose estimation, but their predictions might produce weird fingers, as illustrated in Fig. \ref{fig_weird_hand}. The fingers might violate the naturally kinematic characteristics of the hand. Inspired by this issue, we propose two new objective functions, being Angle and Direction loss, to remain the kinematic characteristics of the predicted hand. Let $U_t=\{u_{ti}:i\in (1,...,P)\}$ denote the set of $P$ knuckles of the ground-truth hand of the $t$-th frame. Let $U'_t=\{u'_{ti}:i\in (1,...,P)\}$ denote the set of $P$ knuckles of the predicted hand of the $t$-th frame. When computing the angle between two knuckles, we build the knuckle vectors $\overrightarrow{u}_{ti}$ and $\overrightarrow{u'}_{ti}$ by assigning the direction to the knuckles, as illustrated in Fig. \ref{fig_angle_direction_loss}. Angle and Direction loss are formulated as follows.

\textbf{Angle loss:}
\begin{equation}
    \mathcal{L}_a = \sum_{t=1}^{T}\sum_{i=1}^{P}\sum_{j=1}^{P}f(i,j)\|\widehat{(\overrightarrow{u}_{ti},\overrightarrow{u}_{tj})}-\widehat{(\overrightarrow{u'}_{ti},\overrightarrow{u'}_{tj})}\|_2
\end{equation}
where $f(i,j)$ is equal to 1 if two knuckles are consecutive. Otherwise, $f(i,j)$ is equal to 0.

\textbf{Direction loss:}
\begin{equation}
    \mathcal{L}_d = \sum_{t=1}^{T}\sum_{i=1}^{P}\widehat{(\overrightarrow{u}_{ti},\overrightarrow{u'}_{ti})}
\end{equation}

In addition to the proposed loss functions, we also use the 3D pose loss and finger length loss as follows.

\textbf{3D pose loss:}
\begin{equation}
    \mathcal{L}_p = \sum_{t=1}^{T}\sum_{i=1}^{N}\|{x}_{ti}-\widehat{x}_{ti}\|_2
\end{equation}
where $x_{ti}$ and $\widehat{x}_{ti}$ are the ground-truth and predicted 3D coordinates of the $i$-th hand joint of the $t$-th frame, respectively.

\textbf{Finger length loss:}
\begin{equation}
    \mathcal{L}_f = \sum_{t=1}^{T}\sum_{i=1}^{L}\|{p}_{ti}-\widehat{p}_{ti}\|_2
\end{equation}
where $p_{ti}$ and $\widehat{p}_{ti}$ are the ground-truth and predicted finger length of the $i$-th finger of the $t$-th frame, respectively.

\textbf{Overall objective function} is defined as:
\begin{equation}
    \mathcal{L} = \lambda_p \mathcal{L}_p + \lambda_f \mathcal{L}_f + \lambda_a \mathcal{L}_a + \lambda_d \mathcal{L}_d
\end{equation}
where $\lambda_p$, $\lambda_f$, $\lambda_a$, and $\lambda_d$ are the loss weights.

\section{Experiments}

We evaluate LG-Hand on First-Person Hand Action Benchmark (FPHAB) dataset \cite{garcia2018first}. Our method is compared with ST-GCN \cite{cai2019exploiting} and SST-GCN \cite{le2021sst}. Notably, we train and evaluate the methods using a same codebase for fair comparisons.

\subsection{Training Details}

FPHAB dataset is a large-scale dataset, consisting more than $100K$ frames and $1175$ action sequences . These $1175$ sequences are divided into $45$ action classes, including interactions with $26$ objects at various places such as office or kitchen. 3D hand joint coordinates are created by using 6 magnetic sensors attached to the hand joints. Each hand has $21$ joints. 
In FPHAB dataset, each action consists of multiple sequences, and each sequence has a different length. We use the $3$-rd sequence for evaluation and the rest for training.

\begin{table*}[h!]
\caption{Comparison of the methods on FPHAB dataset. All results are MPJPE (mm). All methods are implemented using the same codebase.}
\begin{center}
\begin{tabular}{|l||c|c|c|c|c|c|c|c|c|c|c|c||c|}
\hline
 & \textbf{was.} & \textbf{unfo.} & \textbf{ope.} & \textbf{rea.} & \textbf{tea.} & \textbf{put.} & \textbf{lig.} & \textbf{toa.} & \textbf{fli.} & \textbf{clo.} & \textbf{use.} & \textbf{squ.} & \textbf{Avg.} \\
 \hline
 ST-GCN \cite{cai2019exploiting} & 18.21 & 16.44 & 20.09 & 22.88 & 20.22 & 22.60 & 18.39 & 21.61 & 18.60 & 17.05 & 34.63 & 21.96 & 20.25 \\
\hline
SST-GCN \cite{le2021sst} & 17.77 & 15.60 & 17.18 & 20.24 & 19.74 & 22.19 & 17.37 & 21.11 & 18.28 & 16.59 & 34.10 & 20.50 & 19.97 \\
\hline
LG-Hand (\textbf{Ours}) & \textbf{13.51} & \textbf{13.39} & \textbf{16.18} & \textbf{16.41} & \textbf{14.84} & \textbf{20.02} & \textbf{15.06} & \textbf{17.81} & \textbf{15.86} & \textbf{14.96} & \textbf{30.02} & \textbf{15.54} & \textbf{17.25} \\
\hline
\end{tabular}
\label{table_main_result}
\end{center}
\end{table*}

The models are trained for $30$ epochs with a batch size of $256$. We use Adam optimizer. The initial learning rate is set to $1e-3$. The learning rate is decreased by a factor of $0.95$ per epoch. Notably, a learning rate decay of $0.5$ is used after each 10 epochs. We use $3$ frames ($T=3$) in our experiments. For the loss weights, we set: $\lambda_p=1$, $\lambda_f=0.1$, $\lambda_a=0.1$, and $\lambda_d=0.01$. All experiments have been done on the hardware using GeForce GTX 1080 GPU and CUDA 11.0. The evaluation metric is Mean Per Joint Position Error (MPJPE), measuring the average Euclidean distance from prediction to ground-truth joint positions.

\subsection{Main Results}

MPJPE measures the average error between the predicted 3D hand joints and the ground-truth 3D hand joints. The results are shown in Table \ref{table_main_result}. Overall, LG-Hand obtains a best result compared to ST-GCN and SST-GCN. Concretely, LG-Hand reduces the overall MPJPE, which is MPJPE over all actions, by 3.00 and 2.72 mm compared to ST-GCN and SST-GCN, as presented in the last column of Table \ref{table_main_result}. Due to the limited space, we only presents the results of 12 actions as in Table \ref{table_main_result}. The result of each action also indicates that LG-Hand outperforms both ST-GCN and SST-GCN. For example with the ``squeeze paper" (``squ.") action, LG-Hand strongly reduces MPJPE by 6.42 and 4.96 mm compared to ST-GCN and SST-GCN, respectively.

\begin{figure}[h!]
\centerline{\includegraphics[width=1.0\linewidth]{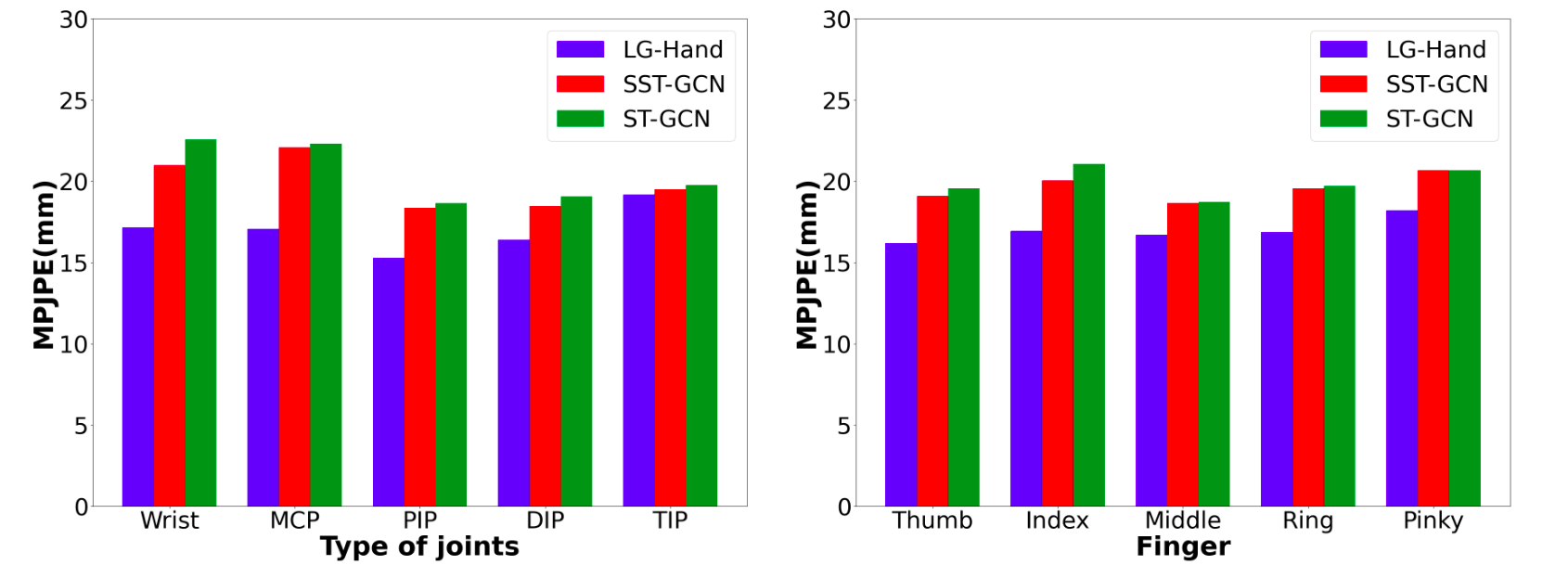}}
\caption{MPJPE (mm) across different parts of the hand.}
\label{fig_joint_finger_error}
\end{figure}

We elaborate the effect of our Angle and Direction loss by measuring MPJPE across different parts of the hand, as illustrated in Fig. \ref{fig_joint_finger_error}. For each method, MPJPEs across the fingers are similar while MPJPEs across the types of joints have a larger gap. This observation is understandable because the fingers play the similar roles while the different types of joints have different connections. For example, wrist connects to 5 MCP-joints of 5 fingers while PIP-joint connects to MCP-joint and DIP-joint on each finger. Especially, the results show that LG-Hand significantly reduces MPJPE of all parts of the hand compared to ST-GCN and SST-GCN.

\subsection{Qualitative Results}

\begin{figure}[h!]
\centerline{\includegraphics[width=1.0\linewidth]{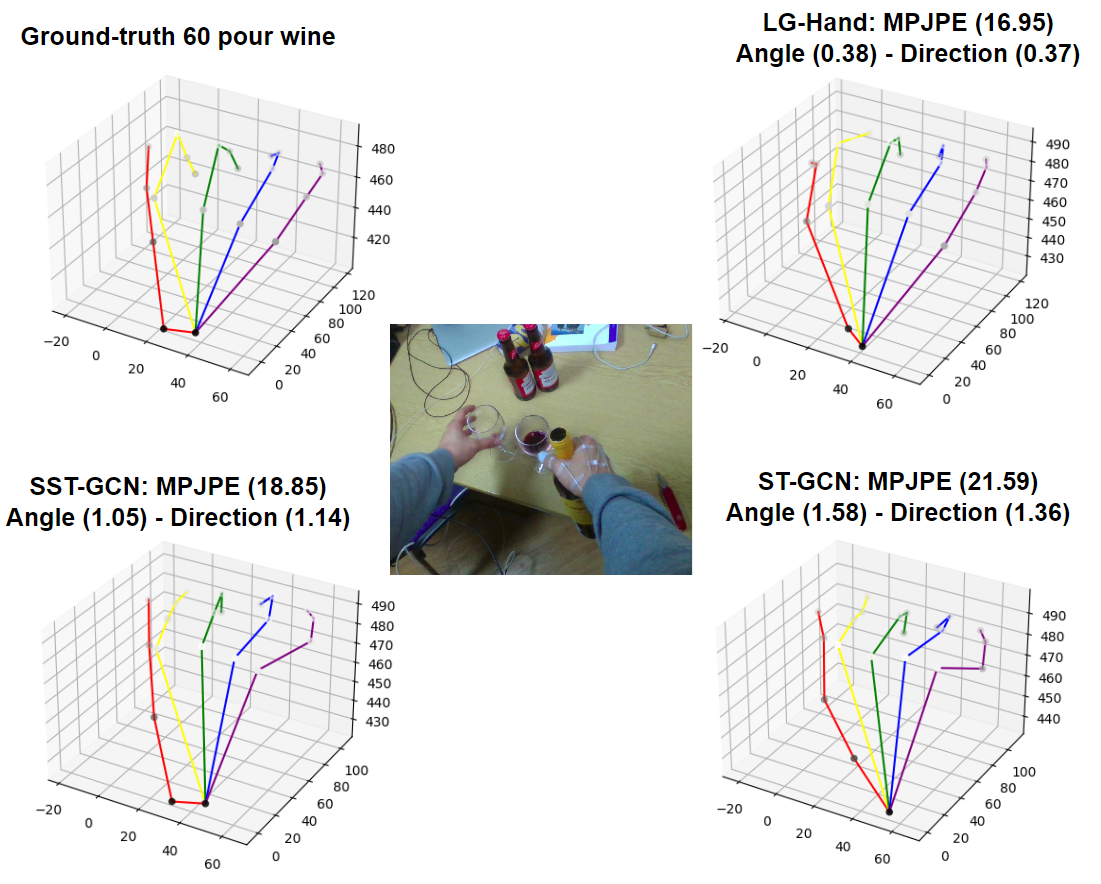}}
\caption{Visualization of the predicted 3D hand joints of the methods for the ``pour wine" action.}
\label{fig_visualization}
\end{figure}

In the sequence of predicted 3D hand joints for the ``pour wine" action, we take the $60$-th frame for visualization in 3D space, as described in Fig. \ref{fig_visualization}.
\begin{itemize}
    \item The results of ST-GCN and SST-GCN show weird predictions such as the ring finger and the pinky finger. These fingers violate the kinematic characteristics of the hand.
    \item By using Angle and Direction loss, LG-Hand is able to produce the better results, where no weird prediction appears.
    \item We also display MPJPE, Angle loss, and Direction loss of the methods as in Fig. \ref{fig_visualization}. LG-Hand obtains the smallest error compared to both ST-GCN and SST-GCN.
\end{itemize}

\begin{figure}[h!]
\centerline{\includegraphics[width=1.0\linewidth]{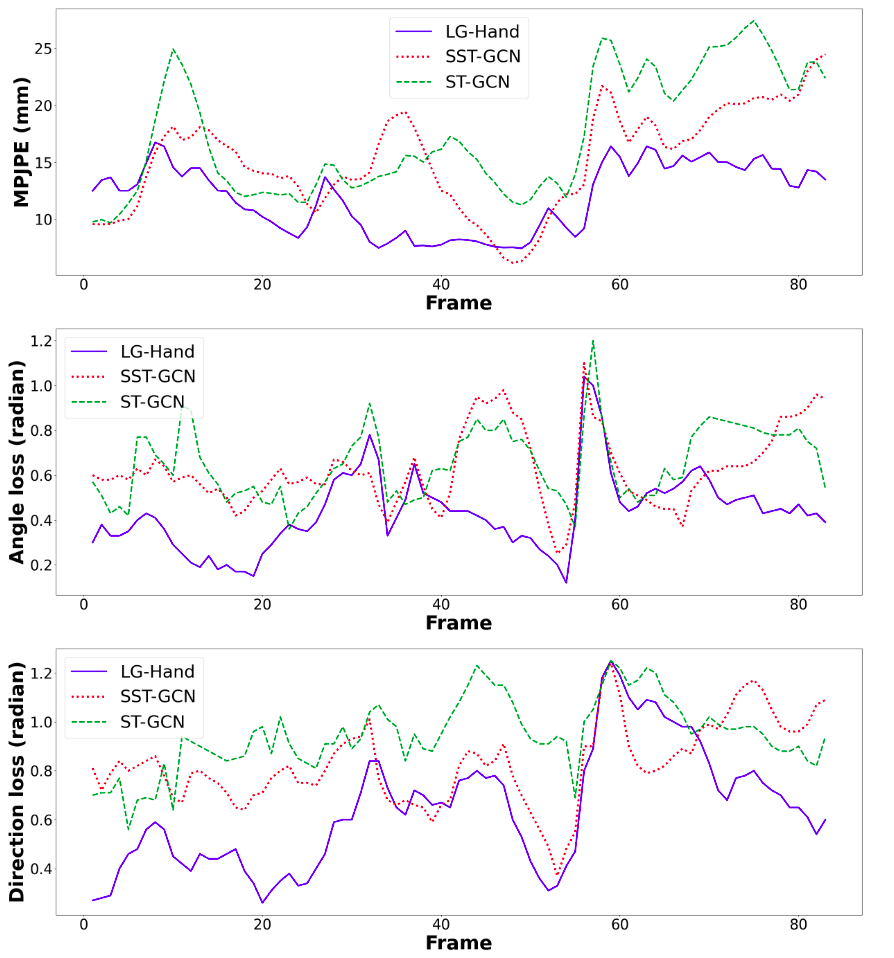}}
\caption{MPJPE, Angle loss, and Direction loss of the methods along a sequence of the ``squeeze paper" action.}
\label{fig_visualization_loss}
\end{figure}

We also examine the tendency of MPJPE, Angle loss, and Direction loss along the sequence of the ``squeeze paper" action, as illustrated in Fig. \ref{fig_visualization_loss}. Compared to ST-GCN and SST-GCN, our LG-Hand has the smaller values of the losses, and our losses tend to strongly decrease along the sequence.

\subsection{Ablation Study}

\subsubsection{Loss weights}
 
We examine the effect of each loss element by varying the loss weights, as shown in Table \ref{table_ablation_loss_weight}. The results are reported on FPHAB dataset with 45 actions. We first train LG-Hand with using only 3D pose loss, and the result is presented in the first row of Table \ref{table_ablation_loss_weight}. We then gradually add Finger length loss, Angle loss, and Direction loss to see how the model behaves. The results show that the model performance is significantly improved by using all four loss functions. For instance, the best setting, the last row of Table \ref{table_ablation_loss_weight}, exhibits a MPJPE improvement of 3.00 mm compared to only using 3D pose loss.

\begin{table}[h!]
\caption{Ablation study with the loss weights.}
\begin{center}
\begin{tabular}{|c|c|c|c||c|}
\hline
\textbf{$\lambda_p$} & \textbf{$\lambda_f$} & \textbf{$\lambda_a$} & \textbf{$\lambda_d$} & \textbf{MPJPE (mm)} \\ 
 \hline
 1 & 0 & 0 & 0 & 20.25 \\
\hline
1 & 0.1 & 0 & 0 & 19.97 \\
\hline
1 & 0.1 & 0.1 & 0 & 18.57 \\
\hline
1 & 0.1 & 0.1 & 0.1 & 18.77 \\
\hline
1 & 0.1 & 0.1 & 0.01 & \textbf{17.25} \\
\hline
\end{tabular}
\label{table_ablation_loss_weight}
\end{center}
\end{table}

\begin{figure}[h!]
\centerline{\includegraphics[width=0.8\linewidth]{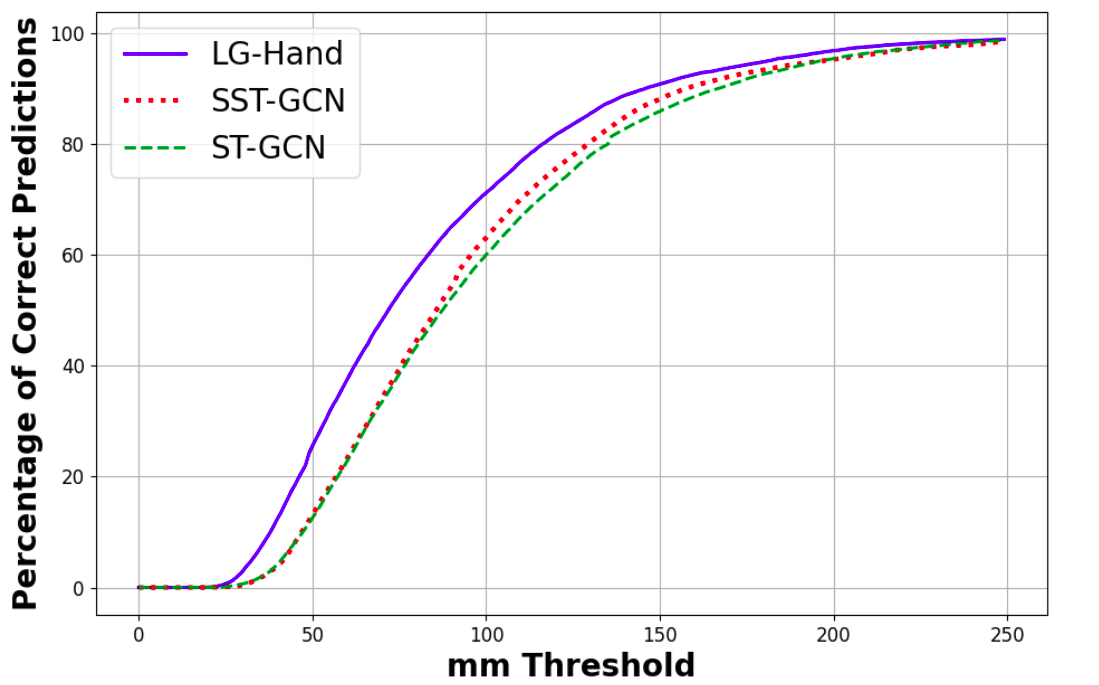}}
\caption{The percentage of correct 3D hand pose on FPHAB dataset.}
\label{fig_pcp}
\end{figure}

\subsubsection{Percentage of correct 3D hand pose}

To elaborate the performance of LG-Hand, we compute the percentage of correct 3D hand poses for various thresholds (measured in millimeters) on FPHAB dataset. A prediction is correct if MPJPE between the ground-truth and predicted 3D hand pose is less than the predefined threshold. As shown in Fig. \ref{fig_pcp}, LG-Hand outperforms SST-GCN and ST-GCN with all thresholds.

\subsubsection{Number of skeleton frames}

\begin{table}[h!]
\caption{Ablation study with the number of skeleton frames.}
\begin{center}
\begin{tabular}{|l||c|c|c|c|c|c|}
\hline
\textbf{\# of frames} & 3 & 5 & 7 & 9 & 11 & 13 \\ 
\hline
\textbf{MPJPE (mm)} & 17.25 & 16.83 & 17.83 & 16.69 & 16.62 & 17.42 \\
\hline
\end{tabular}
\label{table_ablation_n_skeleton_frames}
\end{center}
\end{table}

We examine the behavior of LG-Hand with various amounts of skeleton frames, as shown in Table \ref{table_ablation_n_skeleton_frames}. There is no benefit when increasing the number of skeleton frames. Therefore, LG-Hand uses 3 skeleton frames for a computational efficiency.




\section{Conclusion}

In this paper, we propose LG-Hand, a spatial-temporal GCN-based method for 3D hand pose estimation. We take as input a sequence of consecutive frames of 2D hand joints and output 3D hand joint coordinates. We especially introduce Angle and Direction loss, which can be known as local and global constraint respectively, to incorporate the kinematic information of the hand into the overall objective function. The experimental results show that LG-Hand surpasses the previous methods such as ST-GCN and SST-GCN. For future work, we are interested in researching more objective functions related to the kinematic constraints of the hand to further improve the performance of 3D hand pose estimation.

\bibliographystyle{unsrt}
\bibliography{bibliography}

\begin{thebibliography}{10}

\bibitem{li2019generating}
Chen Li and Gim~Hee Lee.
\newblock Generating multiple hypotheses for 3d human pose estimation with
  mixture density network.
\newblock In {\em Proceedings of the IEEE/CVF Conference on Computer Vision and
  Pattern Recognition}, pages 9887--9895, 2019.

\bibitem{dong2019fast}
Junting Dong, Wen Jiang, Qixing Huang, Hujun Bao, and Xiaowei Zhou.
\newblock Fast and robust multi-person 3d pose estimation from multiple views.
\newblock In {\em Proceedings of the IEEE/CVF Conference on Computer Vision and
  Pattern Recognition}, pages 7792--7801, 2019.

\bibitem{garcia2018first}
Guillermo Garcia-Hernando, Shanxin Yuan, Seungryul Baek, and Tae-Kyun Kim.
\newblock First-person hand action benchmark with rgb-d videos and 3d hand pose
  annotations.
\newblock In {\em Proceedings of the IEEE conference on computer vision and
  pattern recognition}, pages 409--419, 2018.

\bibitem{ge2016robust}
Liuhao Ge, Hui Liang, Junsong Yuan, and Daniel Thalmann.
\newblock Robust 3d hand pose estimation in single depth images: from
  single-view cnn to multi-view cnns.
\newblock In {\em Proceedings of the IEEE conference on computer vision and
  pattern recognition}, pages 3593--3601, 2016.

\bibitem{panteleris2018using}
Paschalis Panteleris, Iason Oikonomidis, and Antonis Argyros.
\newblock Using a single rgb frame for real time 3d hand pose estimation in the
  wild.
\newblock In {\em 2018 IEEE Winter Conference on Applications of Computer
  Vision (WACV)}, pages 436--445. IEEE, 2018.

\bibitem{doosti2020hope}
Bardia Doosti, Shujon Naha, Majid Mirbagheri, and David~J Crandall.
\newblock Hope-net: A graph-based model for hand-object pose estimation.
\newblock In {\em Proceedings of the IEEE/CVF conference on computer vision and
  pattern recognition}, pages 6608--6617, 2020.

\bibitem{cai2019exploiting}
Yujun Cai, Liuhao Ge, Jun Liu, Jianfei Cai, Tat-Jen Cham, Junsong Yuan, and
  Nadia~Magnenat Thalmann.
\newblock Exploiting spatial-temporal relationships for 3d pose estimation via
  graph convolutional networks.
\newblock In {\em Proceedings of the IEEE/CVF International Conference on
  Computer Vision}, pages 2272--2281, 2019.

\bibitem{le2021sst}
Viet-Thanh Le, Thanh-Hai Tran, Van-Nam Hoang, Van-Hung Le, Thi-Lan Le, and Hai
  Vu.
\newblock Sst-gcn: Structure aware spatial-temporal gcn for 3d hand pose
  estimation.
\newblock In {\em 2021 13th International Conference on Knowledge and Systems
  Engineering (KSE)}, pages 1--6. IEEE, 2021.

\bibitem{ge20193d}
Liuhao Ge, Zhou Ren, Yuncheng Li, Zehao Xue, Yingying Wang, Jianfei Cai, and
  Junsong Yuan.
\newblock 3d hand shape and pose estimation from a single rgb image.
\newblock In {\em Proceedings of the IEEE/CVF Conference on Computer Vision and
  Pattern Recognition}, pages 10833--10842, 2019.

\bibitem{martinez2017simple}
Julieta Martinez, Rayat Hossain, Javier Romero, and James~J Little.
\newblock A simple yet effective baseline for 3d human pose estimation.
\newblock In {\em Proceedings of the IEEE international conference on computer
  vision}, pages 2640--2649, 2017.

\bibitem{zimmermann2017learning}
Christian Zimmermann and Thomas Brox.
\newblock Learning to estimate 3d hand pose from single rgb images.
\newblock In {\em Proceedings of the IEEE international conference on computer
  vision}, pages 4903--4911, 2017.

\bibitem{kipf2016semi}
Thomas~N Kipf and Max Welling.
\newblock Semi-supervised classification with graph convolutional networks.
\newblock {\em arXiv preprint arXiv:1609.02907}, 2016.

\end{thebibliography}


\end{document}